\documentclass[5p]{elsarticle}

\usepackage{hyperref,mathtools,mathrsfs}

\usepackage{dsfont}

\usepackage{graphicx}
\usepackage{adjustbox}
\usepackage{enumitem}

\usepackage{url}

\usepackage{bm}

\usepackage{microtype}

\usepackage{amsthm}
\theoremstyle{plain}

\newcommand{\MI}{M\hspace{-1.0mm}I}
\newcommand{\E}{\operatorname{E}}

\newcommand{\bigO}{\ensuremath{\mathcal{O}}}

\newtheorem*{theorem*}{Theorem}
\newtheorem{theorem}{Theorem}
\newtheorem{definition}{Definition}
\newtheorem*{definition*}{Definition}

\begin{document}
	
	\begin{frontmatter}
		
		\title{An entropic feature selection method in perspective of Turing's formula}
		
		
		\author[mymainaddress]{Jingyi Shi}
		
		\author[mysecondaryaddress]{Jialin Zhang}
		
		\author[mymainaddress]{Yaorong Ge}
		
		\address[mymainaddress]{Department of Software and Information Systems, The University of North Carolina at Charlotte}
		\address[mysecondaryaddress]{Department of Mathematics and Statistics, The University of North Carolina at Charlotte}
		
		\begin{abstract}
			Health data are generally complex in type and small in sample size. Such domain-specific challenges make it difficult to capture information reliably and contribute further to the issue of generalization. To assist the analytics of healthcare datasets, we develop a feature selection method based on the concept of Coverage Adjusted Standardized Mutual Information (CASMI). The main advantages of the proposed method are: 1) it selects features more efficiently with the help of an improved entropy estimator, particularly when the sample size is small, and 2) it automatically learns the number of features to be selected based on the information from sample data. Additionally, the proposed method handles feature redundancy from the perspective of joint-distribution. The proposed method focuses on non-ordinal data, while it works with numerical data with an appropriate binning method. A simulation study comparing the proposed method to six widely cited feature selection methods shows that the proposed method performs better when measured by the Information Recovery Ratio, particularly when the sample size is small.
		\end{abstract}
		
		
	\end{frontmatter}
	
	
	\section{Introduction}
	\label{intro}
    Inspired by the recent advancement in Big Data, health informaticians are attempting to assist health care providers and patients from a data perspective, with the hope of improving quality of care, detecting diseases earlier, enhancing decision making, and reducing healthcare costs \cite{kruse2016challenges}. In the process, health informaticians have been confronted with the issue of generalization \cite{lee2017medical}. Analyzing real health data involves many practical problems that could contribute to the issue of generalization; for example, the unknown amount of information (signal) versus error (noise), the curse of dimensionality, and the generalizability of models. All these trivial problems boil down to the essential problem issued by a limited sample. With the limitation of the sample size, the information from the sample cannot represent the information of the population to a desirable extent. For this reason, a simple way to address these trivial problems is to collect a sufficiently large sample, which is unfortunately often impractical in healthcare because of multiple reasons. For example:
    \begin{enumerate}
        \item The term sufficiently large is relative to the dimensionality of data and the complexity of feature spaces. Health data are generally large in dimensionality, particularly when dummy variables (one-hot-encoding) are adopted to represent enormous categories of complex qualitative features (such as extracted words from clinical notes). As a result, a dataset with a sample size of 1,000,000 may not be sufficient, depending on its feature spaces.
        \item There may not be sufficient patient cases for a rare disease. Even if there are ample potential cases, it may be cost-prohibitive for clinical trials to achieve a sufficient sample.  
        
    \end{enumerate}
    
    Without a sufficiently large sample, dimension reduction becomes a major research direction in health data analytic as reducing the dimensionality can partly relieve the issues from a limited sample. These dimension reduction techniques mainly focus on feature selection and feature projection, where feature selection can be further applied to the features created by feature projection. In this article, we focus on feature selection. It has become an important research area, dating back at least to 1997 \cite{blum1997selection}\cite{kohavi1997wrappers}. Since then, many feature selection methods have been proposed and well discussed in multiple recent review papers, such as \cite{li2017feature}, \cite{urbanowicz2018relief}, and \cite{cai2018feature}. To apply these feature selection methods to health data, domain-specific challenges must be considered.
    
    Health data can be numerical and categorical. For example, many machine readings ($e.g.$, heart rate, blood pressure, and blood oxygen level) are numerical, while gene expression data are categorical. A healthcare dataset could contain numerical data only, categorical data only, or a combination of both data types. The fundamental distinction between numerical data and categorical data is whether the data space is ordinal or non-ordinal. As a result, data consisting of only numbers are not necessarily numerical data; for example, gene expression data can be coded to numbers using dummy variables, but it should be still considered as categorical. When the data space is ordinal (numerical data only), classical methods---which detect the association using ordinal information---are more powerful in capturing the associations in data. When the data space is non-ordinal (categorical data only), ordinal information does not naturally exist; hence, continuing to use classical methods onto coded data loses their original advantages and has additional estimation issues. Namely, involving dummy variables increases the dimensionality of data and further exacerbates the estimation problem using a limited sample. This particularly happens when an involved categorical feature has a complex feature space that requires a tremendous number of dummy variables to represent all the different categories. To deal with the categorical data, only information-theoretic quantities ($e.g.$, entropy and mutual information \cite{shannon1948mathematical}) serve the purpose. When a dataset is a combination of both data types, it is inconclusive about whether to use classical or information-theoretic methods. In general, if one believes that the numerical data in the dataset carry more information than the categorical data, then classical methods can be used. If one believes the categorical data carry more information, then information-theoretic methods should be used, and the numerical data should be binned to categorical data. One should be advised that coding categorical data for classical methods increases dimensionality and issues more difficulties in estimation, while binning numerical data for information-theoretic methods inevitably loses ordinal information. It should also be noted that, although ordinal information could provide extra information about associations among the data, the ordinal information could also mislead a person's judgment when associations actually exist, but there is no visual pattern among the data. The way that classical methods work is very similar to our visualization; if there is a pattern that can be visually observed, then it can also be detected by some classical methods. However, not all associations among numerical data are visually observable, in which case, classical methods would fail to detect the associations. On the other hand, if there is a visual pattern among data, binning the data (losing the ordinal information) would not necessarily lead to a loss of associations among data; it depends on the binning methods and performance of the information-theoretic methods.

    Classical feature selection methods include, but are not limited to, Fisher Score \cite{duda2012pattern}, ReliefF \cite{robnik2003theoretical}, Trace Ratio \cite{nie2008trace}, Laplacian Score \cite{he2006laplacian}, SPEC \cite{zhao2007spectral}, $l_p$-regularized \cite{liu2009slep}, $l_{p,q}$-regularized \cite{liu2009slep}, Efficient and Robust Feature Selection (REFS) \cite{nie2010efficient}, Multi-Cluster Feature Selection (MCFS) \cite{cai2010unsupervised}, Unsupervised Feature Selection Algorithm (UDFS) \cite{yang2011robust}, Nonnegative Discriminative Feature Selection (NDFS) \cite{li2012unsupervised}, T-score \cite{davis1986statistics}, and LASSO \cite{tibshirani1996regression}. All these classical feature selection methods require information from ordinal spaces, such as moments ($e.g.$, mean and variance) and spacial information ($e.g.$, nearest location and norms). Information-theoretic feature selection methods include, but are not limit to, Mutual Information Maximisation (MIM) \cite{lewis1992feature}, Mutual Information Feature Selection (MIFS) \cite{battiti1994using}, Joint Mutual Information (JMI) \cite{yang2000data}, minimal Conditional Mutual Information Maximisation (CMIM) \cite{vidal2003object}\cite{fleuret2004fast}, Minimum Redundancy Maximal Relevancy (MRMR) \cite{peng2005feature}, Conditional Infomax Feature Extraction (CIFE) \cite{lin2006conditional}, Informative Fragments (IF) \cite{vidal2003object}, Double Input Symmetrical Relevance (DISR) \cite{meyer2006use}, minimal Normalised Joint Mutual Information Maximisation (NJMIM) \cite{bennasar2015feature}, Chi-square Score \cite{liu1995chi2}, Gini Index \cite{gini1912variabilita}, and CFS \cite{hall1999feature}. All these information-theoretic methods use ordered probabilities, which always exist in non-ordinal spaces. For example, frequencies, category probabilities (proportions), Shannon's entropy, mutual information, and symmetric uncertainty are all functions of ordered probabilities.

    In many cases, all (or most) of the data in a healthcare dataset could be categorical. To analyze the categorical data in such a dataset, information-theoretic feature selection methods are preferred because they could capture the associations among features without using dummy variables, where classical methods require dummy variables that would increase the dimensionality. Most existing information-theoretic methods use entropy or mutual information (a function of entropy) to measure associations among data. Information-theoretic methods that do not use entropy include Gini Index and Chi-square Score. Gini Index focuses on whether a feature is separative, but does not indicate probabilistic associations. Chi-square Score relies on the performance of asymptotic normality on each component, and when there are categories with low frequencies ($e.g.$, less than 5), the Chi-square Score is very unstable. However, under a limited sample, we should expect at least a few, if not many, categories would have relatively low frequencies. For the existing information-theoretic methods that use entropy (we call these \textit{entropic methods}), all of them estimate entropy with the classical maximum likelihood estimator (the plug-in estimator). The plug-in entropy estimator performs very poorly when the sample size is not sufficiently large \cite{harris1975statistical}\cite{paninski2003estimation}, and we have discussed that the sample size is usually relatively limited in healthcare datasets. As a result, to use entropic methods in healthcare data analytics, the estimation of entropy under small samples must be improved.

    In addition to estimation based on small samples, the unhelpful association is another issue with these samples. While the issue of estimation can be addressed by using a better estimator, the problem of unhelpful association is trickier. The unhelpful association is partially a result of sample randomness, and it could be severe when the sample size is small. Suppose there is a healthcare dataset with multiple features and one outcome, and there is a feature in the dataset that could distinguish the values of the outcome based on the sample information, then there are three possible situations:
    \begin{enumerate}[leftmargin=2cm]
        \item[Situation 1] The feature has abundant real information toward the outcome, and the real information is well preserved by the sample data.
        \item[Situation 2] The feature has abundant real information toward the outcome, but the real information is not well preserved by the sample data.
        \item[Situation 3] The feature has little real information but seems relevant to the outcome because of randomness in the sample.
    \end{enumerate}
    The term \textit{real information} of a feature means the feature-carried information that could indicate the values of the outcome at the population level. All three situations are conceptual classifications. At the population level, situation 1 and 2 features are relevant features, and situation 3 features are irrelevant features. It is clear that situation 1 features should be selected while situation 3 features should be dropped. For situation 2, caution should be exercised. Intuitively, situation 2 features should be kept as they are relevant features at the population level. However, as a result of a limited sample, the information carried by these situation 2 features are very subtle. There are at least two constitutional problems about the information from situation 2 features. First, although the feature could distinguish the values of the outcome based on the sample information, the sample-preserved information possibly provides only a meager coverage of all the possible values of the feature. As a result, when there is a new observation ($e.g.$, a new patient), it is very likely that the new observation's corresponding label has not been observed by the preserved information, in which case no outcome information is available to assist prediction based on the information of such a situation 2 feature. Second, because of the limited sample, the predictability of the situation 2 features revealed by the sample may not be complete; hence, it could contribute as an (a) error (noise). For example, based on the sample information, different values of a situation 2 feature could possibly uniquely determine a corresponding value of the outcome (particularly when a feature space is complex while the sample size is small), but this deterministic relationship revealed by a limited sample is unlikely to be true at the population level. As a result, using this information in further modelling and prediction would be wrong and could further contribute to the issue of generalization. Therefore, we suggest omitting situation 2 features. In addition, one should note that a relevant feature being categorized as situation 2 is a consequence of a limited sample. All situation 2 features would eventually become situation 1 when the sample size grows (because more real information would be revealed). As a summary, under a limited sample, situation 1 features should be kept, and situation 2 and 3 features should be dropped.

    Focusing on the domain-specific challenges from health data, we develop the proposed entropic feature selection method based on the concept of Coverage Adjusted Standardized Mutual Information (CASMI). The proposed method aims at improving the performance of estimation and addressing the issue of unhelpful association under relatively small samples. The rest of the article is organized as follows. The concept, intuition, and estimation of CASMI are discussed in Section \ref{CASMIandEE}. The proposed method is described in detail in Section \ref{methods} and evaluated by a simulation study in Section \ref{simulation}. A brief discussion is in Section \ref{discussion}.
	
	\section{CASMI and its Estimation}
	\label{CASMIandEE}
	
	In this section, we introduce the concept, intuition, and estimation of CASMI. Before we proceed, let us state the notations first. 
	
	Let $\mathscr{X}=\{x_i; i=1,\cdots,K_{1}\}$ and $\mathscr{Y}=\{y_j; j=1,\cdots,$ $K_{2}\}$  be two finite alphabets with cardinalities $K_{1}<\infty$ and $K_{2}<\infty$, respectively. Consider the Cartesian product $\mathscr{X\times Y}$ with a joint probability distribution $\mathbf{p}=\{p_{i,j}\}$. Let the two marginal distributions be respectively denoted by $\mathbf{p}_{x}=\{p_{i,\cdot}\}$ and $\mathbf{p}_{y}=\{p_{\cdot,j}\}$, where $p_{i,\cdot}=\sum_{j}p_{i,j}$ and $p_{\cdot,j}=\sum_{i}p_{i,j}$. Assume that $p_{i,\cdot}>0$ and $p_{\cdot,j}>0$ for all $1\leq i \leq K_1$ and $1\leq j \leq K_2$ and that there are $K=\sum_{i,j}1[p_{i,j}>0]$ non-zero entries in $\{p_{i,j}\}$. We re-enumerate these $K$ positive probabilities in one sequence and denote it as $\{p_k; k=1,\cdots, K\}$. Let $X$ and $Y$ be random variables following distributions $\mathbf{p}_{x}$ and $\mathbf{p}_{y}$, respectively. For every pair of $i$ and $j$, let $f_{i,j}$ be the observed frequency of the random pair $(X,Y)$ taking value $(x_{i},y_{j})$, where $i=1,\cdots,K_{1}$ and $j=1,\cdots,K_{2}$, in an $iid$ sample of size $n$ from $\mathscr{X\times Y}$ under $\mathbf{p}$, and let $\hat{p}_{i,j}=f_{i,j}/n$ be the corresponding relative frequency. Consequently, we write $\hat{\mathbf{p}}=\{\hat{p}_{k}\}$ $(i.e., \{\hat{p}_{i,j}\})$,
	$\hat{\mathbf{p}}_{x}=\{\hat{p}_{i,\cdot}\}$, and $\hat{\mathbf{p}}_{y}=\{\hat{p}_{\cdot,j}\}$ as the sets of observed joint and marginal relative frequencies. Shannon's mutual information between $X$ and $Y$ is defined as
	
	\begin{equation}
	\MI(X,Y)=H(X)+H(Y)-H(X,Y),
	\label{MI}
	\end{equation}
	where
	
	\begin{align*}
	H(X)&=-\sum_{i}p_{i,\cdot}\ln p_{i,\cdot}, \\
	H(Y)&=-\sum_{j}p_{\cdot,j}\ln p_{\cdot,j}, \\
	H(X,Y)&=-\sum_{i}\sum_{j}p_{i,j}\ln p_{i,j}=-\sum_{k=1}^K p_k\ln p_k.
	\end{align*}
	
	We define the CASMI as follows:
	
	\begin{definition*}[CASMI]
		$\kappa^*$, the Coverage Adjusted Standardized Mutual Information (CASMI) of a feature $X$ to an outcome $Y$, is defined as
		\begin{equation}
		\kappa ^*(X,Y)=\kappa(X,Y)\cdot(1-\pi_0(X)),
		\end{equation}
		where
	\begin{equation}
	\kappa(X,Y)=\frac{\MI(X,Y)}{H(Y)}, \label{kappa}
	\end{equation}
	and $(1-\pi _0)$ is the sample coverage that was first introduced by Good \cite{good1953population} as ``the proportion of the population represented by (the species occurring in) the sample''.
		\label{CASMI}
	\end{definition*}

	\subsection{Intuition of CASMI}
	
	Many entropic concepts can measure the associations among non-ordinal data; for example, mutual information (MI), Kullback-Leibler divergence (\cite{kullback1951information}), conditional mutual information (\cite{wyner1978definition}), and weighted variants (\cite{guiasu1977information}). Among them, MI is the fundamental concept as all the other entropic association measurements are developed based on or equivalent to MI. For this reason, we develop the CASMI starting with MI. It is well known that $\MI\geq0$, and $\MI(X,Y)=0$ if and only if $X$ and $Y$ are independent. However, MI is not bounded from above; hence, using the values of MI to compare the degrees of dependence among different pairs of random variables is inconvenient. Therefore, it is necessary to standardize the mutual information, which yields to the so-called standardized mutual information (SMI) or normalized variants. \cite{zhang2016statistical} provides several forms of SMI, such as $\MI/H(X)$ (also known as information gain ratio if $X$ is a feature and $Y$ is the outcome), $\MI/H(Y)$, and $\MI/H(X,Y)$. All these forms of SMI can be proven to be bounded by $[0, 1]$, where $0$ stands for independence between $X$ and $Y$, and $1$ stands differently for different SMIs. For $\MI/H(X)$ (information gain ratio), 1 means that, given the value of $Y$ (outcome), the value of $X$ (feature) is determinate. For $\MI/H(Y)$, 1 means that, given the value of $X$, the value of $Y$ is determinate. For $\MI/H(X,Y)$, 1 means a one-to-one correspondence between $X$ and $Y$.
	
	The goal of feature selection is to separate the predictive features from non-predictive features. In this regard, $\MI/H(Y)=1$ is most desirable because $\MI/H(X)=1$ does not indicate the predictability of $X$ and $\MI/H(X,Y)=1$ is too strong and unnecessary. Therefore, we select $\kappa$ in (\ref{kappa}) as the SMI in CASMI.
	
	As we have discussed, detecting unhelpful associations under small samples is important in health data analytics as involving unhelpful associations would bring too much noise or unnecessary dimensions to model-building or prediction. In other words, we would like to detect situation 2 and 3 features in a limited sample. The common characteristics among situation 2 and 3 features is the information revealed by the limited sample covers little of the total information in the population. For this reason, we can use sample coverage ($1-\pi_0$), the concept introduced by Good, to detect these features. A feature with high predictability but low sample coverage must belong to either situation 2 or 3. In CASMI, we multiply the SMI by the sample coverage. Under this setting, although features from situations 2 and 3 have high SMI values, their CASMI scores would be low because of their low sample coverages; hence, these features would not be selected in a greedy selection. On the other hand, the CASMI score for a situation 1 feature would be high because both SMI and the sample coverage are high. As a result, by selecting features greedily, situation 1 features would be selected, while situation 2 and 3 features would be dropped.
	
	The purpose of CASMI is to capture the association between a feature and the outcome, with a penalized term from the sample coverage, so that features under situations 2 and 3 would be eliminated. By selecting features under only situation 1, the issue of generalization under small samples is expected to be reduced. (See Section 3 for a discussion on feature redundancy (or feature interaction).)
	
	It may be interesting to note that the CASMI is an information-theoretic quantity that is related to both the population and the sample. It is neither a parameter nor a statistic, and it is only observable when both the population and the sample are known. Next, we introduce its estimation.
	
	\subsection{Estimation}
	\label{entropyestimation}

	To estimate $\kappa^*$ (CASMI), we need to estimate $\pi _0$ and $\kappa$. $\pi _0 (X)$ can be estimated by Turing's formula \cite{good1953population}
	\begin{equation}
	T_1(X) = N_1(X) / n,
	\label{turingf}
	\end{equation} where $N_1(X)$ is the number of singletons in the sample. For example, if a sample of English letters consists of $\{a,a,a,b,$ $c,c,d,e,e,f\}$, then the corresponding $N_1=3$ ($b, d,$ and $f$ are the three singletons). Discussions on the performance of estimating $\pi_0$ by $T_1$ can be found in \cite{zhang2016statistical} and \cite{ohannessian2012rare}. In experimental categorical data, singletons could possibly indicate the sample size is small. As the sample size grows, the chance of obtaining a singleton in the sample approaches zero. It may be interesting to note that using (\ref{turingf}) to estimate the sample coverage would automatically separates ID-like features. This is because an  ID-like feature is naturally all (or almost all) singletons and would result in a zero (or very small) estimated sample coverage that further leads to a zero (or very low) CASMI score; hence, such an ID-like feature would not be selected.

	Estimating $\kappa(X,Y)$ is equivalent to estimating $\MI(X,Y)$ and $H(Y)$. As we have discussed, thus far, all the existing entropic information-theoretic methods use the plug-in estimator of entropy ($\hat{H}$). However, the plug-in entropy estimator has a huge bias, particularly when sample size is small. \cite{harris1975statistical} showed that the bias of $\hat{H}$ is
	
	\begin{equation*}
	    \E(\hat{H})-H=-\frac{K-1}{2n}+\frac{1}{12n^2}\left(1-\sum_{k=1}^{K}\frac{1}{p_{k}}\right)+\bigO\left(n^{-3}\right),\footnote{We write $f=\bigO(g(n))$ to denote $\limsup_{n\to\infty}\left|f(n)/g(n)\right|<\infty$.}
	\end{equation*}
	where $n$ is the sample size and $K$ is the cardinality of the space on which the probability distribution $\{p_k\}$ lives. Based on the expressions of the bias, it is easy to see that the plug-in estimator underestimates the real entropy, and the bias approaches 0 as $n$ (sample size) approaches infinity, with a rate of $n^{-1}$ (power decay). Because of the power decaying rate, the bias is not small when sample size ($n$) is relatively low.
	
	To improve the estimation under a small sample, we adopt the following $\hat{H}_z$ \cite{zhang2012entropy} as the estimator of $H$:
	\begin{equation}
	\hat{H}_z=\sum_{v=1}^{n-1}\frac{1}{v}\frac{n^{1+v}[n-(1+v)]!}{n!}\sum_{k}\left[\hat{p}_k \prod_{j=0}^{v-1}\left(1-\hat{p}_k-\frac{j}{n}\right)\right].
	\label{hzhat}
	\end{equation}
	
	Compared to the power decaying bias of $\hat{H}$, $\hat{H}_z$ has an exponentially decaying bias
	\begin{equation*}
	\E(\hat{H}_z)-H=\bigO\left(\frac{(1-p_{\wedge})^n}{n}\right),
	\end{equation*}
	where $p_{\wedge}=\min\{p_k>0\}$.

	To help understand the differences between the power decaying bias and exponentially decaying bias, we conduct a simulation. In the simulation, the real underlying distribution is $p_k=k/2001000$, where $k=1,2,\dots, 2000$ ($i.e.$, a triangle distribution). Under this setting, the true entropy $H=7.408005$. To compare the two estimators, we independently generate 10,000 samples following the triangle distribution for each of the six sample size settings ($i.e.$, we generate 60,000 random samples in total). The average values of $\hat{H}$ and $\hat{H}_z$ under different sample sizes are summarized in Table \ref{entropycomparison}.
	
	\begin{table}[h!]
	\caption{Estimation comparison between $\hat{H}$ and $\hat{H}_z$.}
	\begin{center}
    \begin{tabular}{ |c|c|c|c|c|c|c| } 
    \hline
    $n$ & 100 & 300 & 500 & 1000 & 1500 & 2000 \\
    \hline
    avg. of $\hat{H}$ & 4.56 & 5.57 & 6.00 & 6.51 & 6.75 & 6.89\\ 
    \hline
    avg. of $\hat{H}_z$ & 5.11 & 6.09 & 6.49 & 6.92 & 7.11 & 7.21\\ 
    \hline
    \end{tabular}
    \label{entropycomparison}
    \end{center}
	\end{table}
	
	The calculation shows that $\hat{H}$ would consistently underestimate $H$ more than $\hat{H}_z$. The underestimation is more severe when the sample size is smaller. Therefore, from a theoretical perspective, we expect adopting $\hat{H}_z$ in estimating the entropies in CASMI would provide a better estimation, particularly under small samples. Furthermore, we expect CASMI would capture the associations among features and the outcome more accurately under small samples because of the improvement in estimation. Interested readers can find additional discussions on comparison among more entropy estimators in \cite{zhang2012entropy}, and comparison about mutual information estimators using $\hat{H}$ and $\hat{H}_z$ in \cite{zhang2015mutual}.
	
	 Consequently, we let 
	\begin{equation}
	\widehat{\MI}_z (X,Y)=\hat{H}_z (X)+\hat{H}_z (Y)-\hat{H}_z (X,Y),\label{mizhat}
	\end{equation}
	and we estimate $\kappa$ as
	\begin{equation}
	\hat{\kappa}_z(X,Y)=\frac{\widehat{\MI}_z(X,Y)}{\hat{H}_z(Y)}.
	\label{kappazhat}
	\end{equation}
	
	As a summary, we estimate $\kappa^*$ by the following estimator, which is the scoring function of the selection stage in the proposed method.
	\begin{equation}
	\hat{\kappa}^*(X,Y)=\hat{\kappa}_z(X,Y)\cdot (1-T_1(X)),
	\label{scoring}
	\end{equation}
	where $\hat{\kappa}_z$ is defined in (\ref{kappazhat}) and $T_1$ is defined in (\ref{turingf}). $\hat{\kappa}^*$ adopts an entropy estimator with an exponentially decaying bias to improve the performance in estimating $\kappa^*$ and capturing the associations when the sample size is not sufficiently large. Furthermore, we expect involving the sample coverage would separate and drop situation 2 and 3 features under small samples.

	\section{CASMI Based Feature Selection Method} 
	
	\label{methods}
	
	In this section, we introduce the proposed feature selection method in detail. The proposed method contains two stages. Before we present the two stages, let us first discuss data preprocessing.
	
	\subsection*{\ref{methods}.0. Data preprocessing} 
	To use the proposed method, all features and the outcome data must be preprocessed to categorical data. Continuous numerical data must be discretized, and there are numerous discretization methods \cite{dougherty1995supervised}. While binning continuous features, the estimated sample coverage (\ref{turingf}) should be checked to avoid over-discretization, which increases the risk of wrongly shifting a feature from situation 1 to situation 2.
	
	If the data are already categorical, one may need to combine some of the categories to improve the sample coverage, when necessary. When most observations of a feature are singletons, then the coverage is close to 0, in which case it is difficult to draw any reliable and generalizable statistical inference. Therefore, for features that may carry real information but have low sample coverages (below 50\%), it is suggested to regroup them to create repeats and improve coverages. Note that not all features are worth  regrouping; for example, if a feature is the IDs of patients, regrouping should be avoided as there is no reason to believe an ID can contribute to the outcome. The proposed method does not select features with low sample coverages; hence, ID-like features are eliminated automatically.
	
	When a feature contains missing (or invalid) data that cannot be recovered by the data collector, without deleting the feature, there are several possible remedies, such as deleting the observation, making an educated guess, predicting the missing values, and listing all missing values as NA. While it is the user's preference on how to handle the missing data, one should be advised that manipulating (guessing or predicting) the missing data could create (or enhance) false associations; therefore, one should be cautious. Assigning all the missing values as NA generally would not create false associations, but it may reduce the predictive information of the feature. The performance of each remedy method could vary from situation to situation. Additional discussions on handling missing data can be found in \cite{rubin1976inference}, \cite{little2012prevention}, and \cite{kang2013prevention}. We suggest dealing with the missing data at the beginning of the data preprocessing.

	The processed data should contain only categorical features and outcome(s). A feature with only integer values could be considered as categorical as long as the sample coverage is satisfactory.

	\subsection{Stage 1: Eliminate independent features}
	In this stage, we eliminate the features that are believed to be independent of the outcome based on a statistical test. This step filters out the features that are very unlikely to be useful; hence, the computation time for feature selection is reduced.
	
	Suppose there are $p$ features, $X_1, X_2, \dots, X_p$, and one outcome, $Y$, in a dataset. Note that there could be multiple outcome attributes in a dataset. Because each outcome attribute has its own related features, when making a feature selection, we consider one outcome attribute at a time.
	
	In finding independent features, we adopt a chi-squared test of independence using $\widehat{\MI}_z$ as the statistic.
	\begin{theorem}\cite{zhang2018mutual}
		Provided that $\MI=0$,
		\begin{align}
		\chi^2=2n\widehat{\MI}_z+(K_1-1)(K_2-1)\stackrel{L}{\rightarrow}\chi^2\left((K_1-1)(K_2-1)\right),
		\end{align}
		where $\widehat{\MI}_z$ is defined in (\ref{mizhat}). $K_1$ and $K_2$ are the effective cardinalities of the selected feature $X$ and the outcome $Y$, respectively.\footnote{We write $\stackrel{L}{\rightarrow}$ to denote convergence in distribution.}
		\label{testind}
	\end{theorem}
	Compared to Pearson's chi-squared test of independence, testing independence using Theorem \ref{testind} has more statistical power, particularly when the sample size is small \cite{zhang2018mutual}. We test hypothesis $H_0 : \MI(X, Y) =0$ against $H_a : \MI(X,Y)> 0$ between the outcome and each of the features. At a user-chosen level of significance ($\alpha$), any feature whose test decision fails to reject $H_0$ is eliminated at this stage. It is suggested to let $\alpha=0.10$. A smaller $\alpha$ increases the chance of Type-II error (eliminating useful features); a larger $\alpha$ reduces the ability of the elimination, which results in a longer selection computation time in the next stage.

	Let $X_{1}, X_{2}, \dots, X_{s}$ denote the $s$ features (out of the $p$ features) that have passed the test of independence. The other $(p-s)$ features are eliminated at this stage. Note that the $X_1, \dots, X_s$ are temporary notations for features. Namely, the $X_1$ in $\{X_1,\dots, X_p\}:=\{X\}_p$ and the $X_1$ in $\{X_1, \dots, X_s\}:=\{X\}_s$ are different if the $X_1$ in $\{X\}_p$ is eliminated in this stage. Note that we do not consider feature redundancy at Stage 1. Redundant features could all pass the test of independence as long as they appear to be relevant to the outcome based on sample data. Feature redundancy would be considered at Stage 2.
	
	\subsection{Stage 2: Selection}
	
	In this stage, we make a greedy selection among the $s$ remaining features from Stage 1.
	
	The selection algorithm is:
	
	\begin{enumerate}[leftmargin=4.5mm]
		\item $X_{(1)}=\arg_{X_i \in \{X\}_s} \max \left[\hat{\kappa}^*(X_i,Y)\right]$;\\
		\item $X_{(2)}=\arg_{X_i \in \{X\}_s\backslash\{X_{(1)}\}} \max \left[\hat{\kappa}^*(X_{(1)}\times X_i ,Y)\right]$;\\
		\item $X_{(3)}=\arg_{X_i \in \{X\}_s\backslash\{X_{(1)}, X_{(2)}\}} \max [\hat{\kappa}^*(X_{(1)}\times X_{(2)} \times X_i,$\\$~~~~~~~~~~~Y)]$;\\
		\\$\cdots$\\
		The algorithm stops at time $c$ when $\hat{\kappa}^*(X_{(1)}\times \cdots \times X_{(c+1)} ,Y)<\hat{\kappa}^*(X_{(1)}\times \cdots \times X_{(c)} ,Y)$.
	\end{enumerate}
	
	To clarify the notations, $\hat{\kappa}^*(X_{(1)}\times X_i ,Y)$ stands for the estimated CASMI of the joint feature $X_{(1)} \times X_i$ to the outcome $Y$, and $\{X\}_s$ is the collection of the $s$ remaining features.
	
	The proposed method handles feature redundancy by considering joint-distributions among features. Taking $X_{(1)}$ and $X_{(2)}$ as examples, the first step yields the feature $X_{(1)}$, which is the most relevant feature (measured by the estimated CASMI) to the outcome. In the second step, we joint the selected $X_{(1)}$ with each of the remaining $(s-1)$ features, and we evaluate the estimated CASMIs between each of the joint-features and the outcome. The joint-feature with the highest estimated CASMI is selected, which becomes $X_{(2)}$. It should be noted that $X_{(1)}$ and $X_{(2)}$ are neither necessarily independent nor necessarily the least dependent. Selecting $X_{(2)}$ only indicates that based on the information provided from $X_{(1)}$, $X_{(2)}$ provides the most additional information about the outcome among the remaining $(s-1)$ features. In addition, CASMI is an information-theoretic quantity that does not use ordinal information of features; therefore, both linear and nonlinear redundancy are captured, evaluated, and considered.
	
	The proposed algorithm stops when the term $\max [\hat{\kappa}^* (\cdot, Y)]$ starts to decrease. The features selected by the proposed method are $X_{(1)},\dots, X_{(c)}$. 
	
	In some situations, a researcher may want to select a desired number of features ($d$) that is different from $c$. For example, let $c=10$, $d_1=6$, and $d_2=15$. When $c=10$ and $d_1=6$, because $6 \leq 10$, we can stop the algorithm at the time 6. When $c=10$ and $d_2=15$, because $15>10$, the user needs to select 5 additional features. We propose two choices on how to select the additional features.
	
	\begin{enumerate}[leftmargin=2cm]
		\item [Choice 1.] Keep running the proposed algorithm until time 15.
		\item [Choice 2.] Use any other user-preferred feature selection methods to select the 5 additional features.
	\end{enumerate}
	
	Choice 2 could be complicated. If the user-preferred feature selection method has a ranking on the selected features, such as filter methods, then one can find the additional features by looking for the top 5 features other than the already-selected 10 features. If the user-preferred feature selection method does not have a ranking among the selected features, one can start by selecting 15 features using the preferred method, and then check if there are exactly 5 new features in the group compared to the 10 features selected by the proposed method. If the number of new features in the group is more than 5, then one needs to reduce the number of selected features, using the preferred method, until a point that there are exactly 5 new features in the group, so that the 5 additional features can be determined.
	
	After the two stages, the proposed method is completed. The performance of the proposed method is evaluated in the following section.

	\section{Simulations}
	
	\label{simulation}
	
	In this section, we provide a simulation study to evaluate the performance of the proposed feature selection method. We first discuss the evaluation metric and then introduce the simulation setup and results.
	
	\subsection{Evaluation Metric}

	The proposed feature selection method selects only relevant features but does not provide an associated model or classifier. In evaluating such a feature selection method, there are two possible approaches \cite{pascoal2017theoretical}. The first approach is to embed a classifier and compare the accuracy of the classification process based on a real dataset. The results obtained with this approach are difficult to generalize as they depend on the specific classifier used in the comparison. The second approach is based on a scenario defined by an initial set of features and a relation between these features and the outcome. Under this situation, a feature selection method could be evaluated by the truth. Focusing on the evaluation of the selected features, we adopt the second approach to evaluate the proposed feature selection method based on the truth. Under this approach, there are several strategies. One can calculate the percentage (success rate) of all relevant features that are selected. For example, let us consider an outcome $T$ that is relevant to three features $F_1$, $F_2$, and $F_3$, where $F_1$ contributes the most information (variability) of $T$, $F_2$ contributes the second most, and $F_3$ contributes the least. Also, there is an irrelevant feature $F_4$ in the dataset. Suppose there are four different selection results: $S_1=\{F_1\}$, $S_2=\{F_1, F_4\}$, $S_3=\{F_2, F_4\}$, and $S_4=\{F_3, F_4\}$. Evaluating their performances using the success rate would achieve the same result (33.3\% or 1/3) for all of them as they all identify one correct feature out of the three. The success rate is simple to calculate because the ground truth is known, and it works well when we focus on the number of correctly selected features or if we assume all the relevant features contribute evenly to the outcome. However, under the restriction of a limited sample, it may be more important to select the group of features that could jointly and efficiently provide the most information instead of selecting all relevant features regardless of the degrees of relevance and redundancy. Although ignoring low relevant or vastly redundant features may lose information, dropping them would further reduce the dimensionality and benefit the estimation. This can be considered as a trade off between estimation (dimensionality) and information: the more information, the more difficult the estimation. When the estimation is overly difficult, the results could be biased and hardly generalizable.

	Because the success rate does not take the degrees of relevance and redundancy into consideration, we introduce the following evaluation metric to measure the ratio of the relevant information from the joint of selected features to the total relevant information from the joint of all the relevant features using mutual information.

	\begin{definition}[Information Recovery Ratio (IRR)]
		
		\label{IRR}
		
		\begin{equation}
		IRR = \frac{\MI (\mathcal{X}_{selected}, Y)}{\MI (\mathcal{X}_{relevant},Y)},
		\label{IRR}
		\end{equation} 
		where $\mathcal{X}_{selected}$ is the random variable that follows the joint-distribution of the selected features, and $\mathcal{X}_{relevant}$ is the random variable that follows the joint-distribution of all the features on which $Y$ depends.
		
	\end{definition}

	The IRR is not calculable in real datasets because 1) there is no knowledge on which features are relevant to the outcome, and 2) the true underlying distributions and associations (including redundancy) of the features and outcomes in real data are unknown. Given the setup of a simulation, we have all the knowledge; hence, the IRR for any group of selected features is calculable.

	The IRR represents the percentage of relevant information in the joint of selected features. It considers feature redundancy by evaluating the mutual information between the joint-feature and the outcome. The range of the IRR is $[0, 1]$. If no relevant features are selected, the IRR is 0. If all the features in the dataset are selected regardless of relevance, the IRR is 1 for certain; therefore, when comparing the performance using the IRR, the number of selected features must be controlled. When the number of selected features from different methods are the same, a larger IRR means the joint of the selected features contains more relevant information; hence, the method is more efficient in dimension reduction. The efficiency of a feature selection method is desirable, particularly under small samples.

	To make a comparison between the IRR and the success rate, both evaluate the performance of feature selection methods only when the ground truth is known. The success rate focuses on the ratio of the number of relevant features selected to the total number of relevant features, while the IRR focuses on the ratio of the relevant information in the joint of the selected features to the total relevant information.

	\subsection{Simulation Setup}
	
	A good evaluation scenario must include a representative set of features, containing relevant, redundant, and irrelevant ones \cite{pascoal2017theoretical}. In the simulation, we generate ten $X$ variables ($X_1,\dots,$ $ X_{10}$) and one outcome ($Y$). Among these variables, $X_1, X_2, X_3, X_4$ (or $X_6$), and $X_5$ are relevant features; $X_6$ (or $X_4$) is a redundant feature; $X_7, X_8, X_9,$ and $X_{10}$ are irrelevant features. The detailed settings are as follows.
	\begin{equation}
	Y = X_1 + X_2 + X_3^3 - 0.5 \cdot X_4^2 + |X_5| + X_6 + \varepsilon,
	\label{simulationModel}
	\end{equation} 
	
	where
	
	\begin{align*}
	X_1=&-3.5\cdot \mathds{1}[Z_1 <-3]-1.4\cdot\mathds{1}[-3 \leq Z_1 \leq -0.5]\\&+\mathds{1}[0.5\leq Z_1 \leq 3] +2.2\cdot \mathds{1}[Z_1 >3],\\
	X_2=&-5\cdot \mathds{1} [Pois_1=0]-3\cdot \mathds{1} [Pois_1=1] \\&+2.4\cdot \mathds{1} [Pois_1=3 \text{ or } 4]+5.4\cdot \mathds{1}[Pois_1 \geq 5],\\
	X_3=&-2 \cdot \mathds{1}[U_1 \leq -0.6]- \mathds{1}[-0.6<U_1 <-0.2]\\
	&+\mathds{1}[0.2<U_1 <0.6]+2\cdot \mathds{1}[U_1 \geq 0.6],\\
	X_4=&(B_1-2)\cdot\mathds{1}[B_1 \neq 4]+5\cdot \mathds{1}[B_1 =4],\\
	X_5=&-2.5\cdot \mathds{1}[Z_2 <-0.5]-2\cdot \mathds{1}[-0.5 \leq Z_2 \leq -0.2]\\&+1.7 \cdot \mathds{1}[-0.2\leq Z_2 \leq 0.2]+2\cdot \mathds{1}[0.2\leq Z_2 \leq 0.6]\\ &+4\cdot \mathds{1}[Z_2 >0.6],\\
	X_6=&X_4,\\
	X_7=&(Pois_2 -2)\cdot \mathds{1} [Pois_2<2]+2\cdot \mathds{1}[Pois_2 \geq 2],\\
	X_8=&-2 \cdot \mathds{1}[U_2 \leq -0.6]- \mathds{1}[-0.6<U_2 <-0.2]\\
	&+\mathds{1}[0.2<U_2 <0.6]+2\cdot \mathds{1}[U_2 \geq 0.6],\\
	X_9=&B_2-1.2,\\
	X_{10}=&-2\cdot \mathds{1}[Z_3 <-1.5]-1.5 \cdot \mathds{1}[-1.5 \leq Z_3 \leq -0.7]\\&+1.5 \cdot \mathds{1}[0.7\leq Z_3 \leq 1.5] +2\cdot \mathds{1}[Z_3 >1.5],\\
	\varepsilon=&-\mathds{1}[U_3\leq \frac{1}{3}]+\mathds{1}[U_3 \geq \frac{2}{3}],
	\end{align*}
	and
	\begin{align*}
	Z_1, Z_2, Z_3 \sim & N(0,1),\\
	Pois_1, Pois_2 \sim & Poisson(2),\\
	B_1 \sim & Binomial(4,0.1),\\
	B_2 \sim & Binomial(6,0.2),\\
	U_1, U_2 \sim & Uniform(-1,1),\\
	U_3\sim & Uniform(0,1).
	\end{align*}

	Usually, a simulation setup should include varieties to justify the challenges in real world data. Namely, it is often desirable to have complex feature spaces and complicated relationships among the features and the outcome. However, the above simulation setup is not complicated for the following reasons.
	
	\begin{enumerate}
	    \item The purpose of this simulation is to evaluate the performance of the proposed method, particularly when the sample size is relatively small. The complexity of the feature spaces and the relationships among the features and the outcome would determine the threshold of what constitutes a sufficiently large sample. As they are not complex, we sample with smaller sizes to evaluate the performances in simulation. This is fair to all feature selection methods in comparison as they select features based on the same sample data with the same sample size.
	    \item The proposed feature selection method is one of the entropic methods. In the simulation, we would compare the performance of the proposed method to only other entropic methods because of the domain-specific challenges discussed in Section \ref{intro}. During the simulation, we assign numerical values to the $X$ variables so that we can generate the value of the outcome $Y$ based on a model. But entropic methods do not use the ordinal information from the numerical data as the inputs of the entropic methods are the frequencies of different numbers. Therefore, involving a complicated model (linear or nonlinear) does not affect the entropic methods because they regard the numbers as labels without ordinal information. However, complicating the model could make the outcome variable $Y$ more complex and result in a higher threshold of a sufficiently large sample, which does not affect the comparison and evaluation among different methods, as discussed previously.
	    \item In calculating IRR, we need the two joint-distributions, $\mathcal{X}_{selected}\times Y$ and $\mathcal{X}_{relevant}\times Y$. To obtain the true joint-distributions, we have to enumerate the combinations among all possible values of the selected relevant features and of all the relevant features with their probabilities, respectively. Complicating the relevant $X$ variables would make the calculation of the joint-distributions unnecessarily complex.
	\end{enumerate}
	
	Note that the major benefit of a simple simulation setup is the ease in calculating the true joint-distributions, which are components of the IRR. In real world data, we do not need such calculations as the true joint-distributions and the IRR are not calculable. Hence, when applying the proposed method on real world high dimensional and complex data, the main calculation is just the estimated CASMI, which is not a problem.
	
	With the simulation setup, one can consider that we create a dataset for evaluation. In this case, we know the ground truth that the features $X_1, X_2, X_3, X_4$ (or $X_6$), and $X_5$ should be selected. We would evaluate the performances by calculating the IRRs for features selected by different methods.
	
	\subsection{Simulation Results}
	In the simulation, we compare the IRR of the proposed feature selection method to the IRRs of six widely cited entropic feature selection methods: MIM, JMI, CMIM, MRMR, DISR, and NJMIM. 
	
	These six entropic methods all require users to set the number of features to be selected, while the proposed method can automatically decide the most appropriate number of features based on data. As we must control the number of selected features to validate the comparison of IRRs,  we use the number of selected features from the proposed method as the number of features to be selected in the six entropic methods in each iteration. It should be noted that we are not claiming the number of features determined by the proposed method is correct. We set them to be the same only for the purpose of validating the comparison. As a matter of fact, the relevant features would not be entirely selected until the sample size is sufficiently large, and the threshold of a sufficiently large sample varies from method to method.

\begin{figure}[h!]
		\includegraphics[width=\columnwidth]{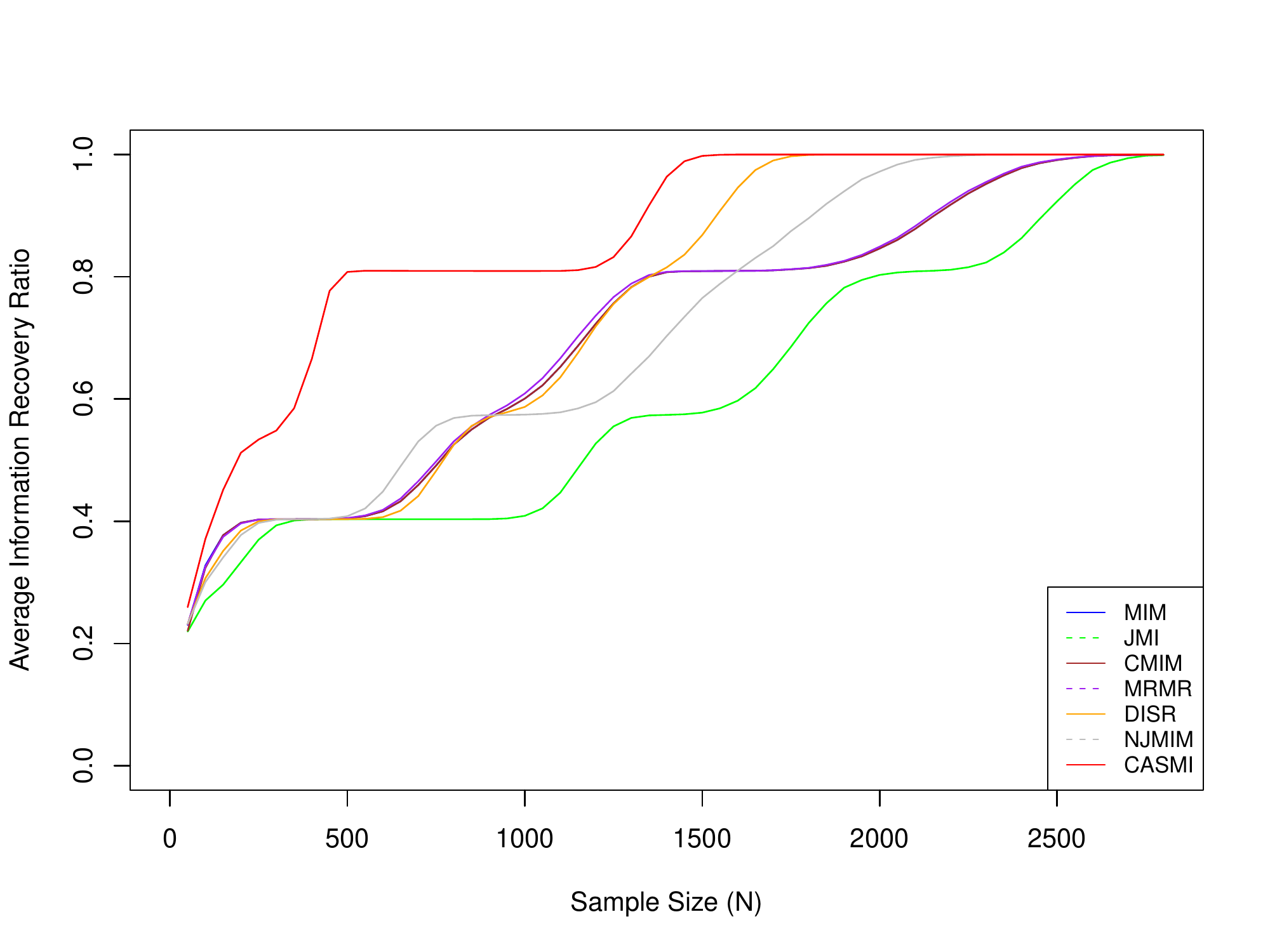}
		\caption{The average IRRs for seven methods, where CASMI refers to the proposed method. The proposed method is the most efficient method when sample size is limited. In the simulation, the threshold of a sufficiently large sample for the proposed method is approximately $N=1500$, which is the smallest among all methods. The vertical index is the IRR, not the success rate. An IRR of 0.8 means 80\% of the total mutual information has been accounted for by the selected features. It does not mean 80\% of relevant features are selected. The proposed method does not select all relevant features when the sample size is small because some relevant features are in situation 2 under a limited sample; hence, they are not selected. As the sample size grows, all situation 2 features eventually become situation 1 features.}
		\label{figure1}
	\end{figure}

	For each sample size $N$ in $\{50,100,150,\dots,2750,2800\}$, we re-generate the entire dataset 10,000 times and calculate the average IRRs of each method. The average IRR results are plotted in Figure \ref{figure1}. 
	
	Based on the results, we can see that the average IRR of the proposed method is consistently higher than or equivalent to all the other methods. This is because under the restriction of a limited sample, the proposed method has a much smaller estimation bias so that it captures the associations among features and the outcome more accurately than the existing methods that estimate with the plug-in estimators. Table \ref{ci} presents the 95\% confidence intervals for IRRs based on features selected by different methods under different sample sizes. Based on the table, we can roughly rank the proposed methods and the six methods as follows: CASMI $>$ DISR $>$ NJMIM $>$ MRMR $>$ MIM $\sim$ CMIM $>$ JMI.

	\begin{table*}[t]
	\caption{The 95\% Confidence Intervals for IRRs based on features selected by different methods under different sample sizes. }
	\centering
\begin{tabular}{cccccccc}
n    & CASMI CI                                   & MIM CI                                     & JMI CI                                     & CMIM CI                                    & MRMR CI                                    & DISR CI                                    & NJMIM CI                                   \\
\hline
50   & {[}0, 0.40{]}                 & {[}0, 0.40{]}                 & {[}0, 0.40{]}                 & {[}0, 0.40{]}                 & {[}0, 0.40{]}                 & {[}0, 0.40{]}                 & {[}0, 0.40{]}                 \\
100  & {[}0.21, 0.57{]}  & {[}0.21, 0.40{]} & {[}0.21, 0.40{]} & {[}0.21, 0.40{]} & {[}0.21, 0.40{]} & {[}0.21, 0.40{]} & {[}0.21, 0.40{]} \\
150  & {[}0.26, 0.62{]}  & {[}0.21, 0.40{]} & {[}0.21, 0.40{]} & {[}0.21, 0.40{]} & {[}0.21, 0.40{]} & {[}0.21, 0.40{]} & {[}0.21, 0.40{]} \\
200  & {[}0.32, 0.62{]}  & {[}0.21, 0.40{]} & {[}0.21, 0.40{]} & {[}0.21, 0.40{]} & {[}0.21, 0.40{]} & {[}0.21, 0.40{]} & {[}0.21, 0.40{]} \\
250  & {[}0.37, 0.62{]}   & {[}0.40, 0.40{]} & {[}0.21, 0.40{]} & {[}0.40, 0.40{]} & {[}0.40, 0.40{]} & {[}0.40, 0.40{]} & {[}0.21, 0.40{]} \\
300  & {[}0.37, 0.62{]}   & {[}0.40, 0.40{]} & {[}0.21, 0.40{]} & {[}0.40, 0.40{]} & {[}0.40, 0.40{]} & {[}0.40, 0.40{]} & {[}0.40, 0.40{]} \\
350  & {[}0.37, 0.81{]}  & {[}0.40, 0.40{]} & {[}0.40, 0.40{]} & {[}0.40, 0.40{]} & {[}0.40, 0.40{]} & {[}0.40, 0.40{]} & {[}0.40, 0.40{]} \\
400  & {[}0.37, 0.81{]}  & {[}0.40, 0.40{]} & {[}0.40, 0.40{]} & {[}0.40, 0.40{]} & {[}0.40, 0.40{]} & {[}0.40, 0.40{]} & {[}0.40, 0.40{]} \\
450  & {[}0.62, 0.81{]}  & {[}0.40, 0.40{]} & {[}0.40, 0.40{]} & {[}0.40, 0.40{]} & {[}0.40, 0.40{]} & {[}0.40, 0.40{]} & {[}0.40, 0.40{]} \\
500  & {[}0.81, 0.81{]} & {[}0.40, 0.40{]} & {[}0.40, 0.40{]} & {[}0.40, 0.40{]} & {[}0.40, 0.40{]} & {[}0.40, 0.40{]} & {[}0.40, 0.57{]}  \\
550  & {[}0.81, 0.81{]} & {[}0.40, 0.57{]}  & {[}0.40, 0.40{]} & {[}0.40, 0.57{]}  & {[}0.40, 0.57{]}  & {[}0.40, 0.40{]} & {[}0.40, 0.57{]}  \\
600  & {[}0.81, 0.81{]} & {[}0.40, 0.57{]}  & {[}0.40, 0.40{]} & {[}0.40, 0.57{]}  & {[}0.40, 0.57{]}  & {[}0.40, 0.40{]} & {[}0.40, 0.57{]}  \\
650  & {[}0.81, 0.81{]} & {[}0.40, 0.57{]}  & {[}0.40, 0.40{]} & {[}0.40, 0.57{]}  & {[}0.40, 0.57{]}  & {[}0.40, 0.57{]}  & {[}0.40, 0.57{]}  \\
700  & {[}0.81, 0.81{]} & {[}0.40, 0.57{]}  & {[}0.40, 0.40{]} & {[}0.40, 0.57{]}  & {[}0.40, 0.57{]}  & {[}0.40, 0.57{]}  & {[}0.40, 0.57{]}  \\
750  & {[}0.81, 0.81{]} & {[}0.40, 0.57{]}  & {[}0.40, 0.40{]} & {[}0.40, 0.57{]}  & {[}0.40, 0.57{]}  & {[}0.40, 0.57{]}  & {[}0.40, 0.57{]}  \\
800  & {[}0.81, 0.81{]} & {[}0.40, 0.57{]}  & {[}0.40, 0.40{]} & {[}0.40, 0.57{]}  & {[}0.40, 0.57{]}  & {[}0.40, 0.57{]}  & {[}0.40, 0.57{]}  \\
850  & {[}0.81, 0.81{]} & {[}0.40, 0.57{]}  & {[}0.40, 0.40{]} & {[}0.40, 0.57{]}  & {[}0.40, 0.57{]}  & {[}0.40, 0.57{]}  & {[}0.57, 0.57{]}   \\
900  & {[}0.81, 0.81{]} & {[}0.40, 0.62{]}  & {[}0.40, 0.40{]} & {[}0.40, 0.62{]}  & {[}0.40, 0.81{]} & {[}0.40, 0.57{]}  & {[}0.57, 0.57{]}   \\
950  & {[}0.81, 0.81{]} & {[}0.57, 0.81{]}  & {[}0.40, 0.40{]} & {[}0.57, 0.81{]}  & {[}0.57, 0.81{]}  & {[}0.57, 0.57{]}   & {[}0.57, 0.57{]}   \\
1000 & {[}0.81, 0.81{]} & {[}0.57, 0.81{]}  & {[}0.40, 0.57{]}  & {[}0.57, 0.81{]}  & {[}0.57, 0.81{]}  & {[}0.57, 0.81{]}  & {[}0.57, 0.57{]}   \\
1050 & {[}0.81, 0.81{]} & {[}0.57, 0.81{]}  & {[}0.40, 0.57{]}  & {[}0.57, 0.81{]}  & {[}0.57, 0.81{]}  & {[}0.57, 0.81{]}  & {[}0.57, 0.57{]}   \\
1100 & {[}0.81, 0.81{]} & {[}0.57, 0.81{]}  & {[}0.40, 0.57{]}  & {[}0.57, 0.81{]}  & {[}0.57, 0.81{]}  & {[}0.57, 0.81{]}  & {[}0.57, 0.57{]}   \\
1150 & {[}0.81, 0.81{]} & {[}0.57, 0.81{]}  & {[}0.40, 0.57{]}  & {[}0.57, 0.81{]}  & {[}0.57, 0.81{]}  & {[}0.57, 0.81{]}  & {[}0.57, 0.81{]}  \\
1200 & {[}0.81, 1{]}                 & {[}0.57, 0.81{]}  & {[}0.40, 0.57{]}  & {[}0.57, 0.81{]}  & {[}0.57, 0.81{]}  & {[}0.57, 0.81{]}  & {[}0.57, 0.81{]}  \\
1250 & {[}0.81, 1{]}                 & {[}0.57, 0.81{]}  & {[}0.40, 0.57{]}  & {[}0.57, 0.81{]}  & {[}0.57, 0.81{]}  & {[}0.57, 0.81{]}  & {[}0.57, 0.81{]}  \\
1300 & {[}0.81, 1{]}                 & {[}0.57, 0.81{]}  & {[}0.40, 0.57{]}  & {[}0.57, 0.81{]}  & {[}0.57, 0.81{]}  & {[}0.57, 0.81{]}  & {[}0.57, 0.81{]}  \\
1350 & {[}0.81, 1{]}                 & {[}0.57, 0.81{]}  & {[}0.57, 0.57{]}   & {[}0.57, 0.81{]}  & {[}0.57, 0.81{]}  & {[}0.57, 0.81{]}  & {[}0.57, 0.81{]}  \\
1400 & {[}0.81, 1{]}                 & {[}0.81, 0.81{]} & {[}0.57, 0.57{]}   & {[}0.81, 0.81{]} & {[}0.81, 0.81{]} & {[}0.81, 1{]}                 & {[}0.57, 0.81{]}  \\
1450 & {[}0.81, 1{]}                 & {[}0.81, 0.81{]} & {[}0.57, 0.57{]}   & {[}0.81, 0.81{]} & {[}0.81, 0.81{]} & {[}0.81, 1{]}                 & {[}0.57, 0.81{]}  \\
1500 & {[}1, 1{]}                                 & {[}0.81, 0.81{]} & {[}0.57, 0.57{]}   & {[}0.81, 0.81{]} & {[}0.81, 0.81{]} & {[}0.81, 1{]}                 & {[}0.57, 0.81{]}  \\
1550 & {[}1, 1{]}                                 & {[}0.81, 0.81{]} & {[}0.57, 0.81{]}  & {[}0.81, 0.81{]} & {[}0.81, 0.81{]} & {[}0.81, 1{]}                 & {[}0.57, 1{]}                  \\
1600 & {[}1, 1{]}                                 & {[}0.81, 0.81{]} & {[}0.57, 0.81{]}  & {[}0.81, 0.81{]} & {[}0.81, 0.81{]} & {[}0.81, 1{]}                 & {[}0.57, 1{]}                  \\
1650 & {[}1, 1{]}                                 & {[}0.81, 0.81{]} & {[}0.57, 0.81{]}  & {[}0.81, 0.81{]} & {[}0.81, 0.81{]} & {[}0.81, 1{]}                 & {[}0.57, 1{]}                  \\
1700 & {[}1, 1{]}                                 & {[}0.81, 0.81{]} & {[}0.57, 0.81{]}  & {[}0.81, 0.81{]} & {[}0.81, 0.81{]} & {[}0.81, 1{]}                 & {[}0.81, 1{]}                 \\
1750 & {[}1, 1{]}                                 & {[}0.81, 0.81{]} & {[}0.57, 0.81{]}  & {[}0.81, 0.81{]} & {[}0.81, 0.81{]} & {[}1, 1{]}                                 & {[}0.81, 1{]}                 \\
1800 & {[}1, 1{]}                                 & {[}0.81, 0.81{]} & {[}0.57, 0.81{]}  & {[}0.81, 0.81{]} & {[}0.81, 1{]}                 & {[}1, 1{]}                                 & {[}0.81, 1{]}                 \\
1850 & {[}1, 1{]}                                 & {[}0.81, 1{]}                 & {[}0.57, 0.81{]}  & {[}0.81, 1{]}                 & {[}0.81, 1{]}                 & {[}1, 1{]}                                 & {[}0.81, 1{]}                 \\
1900 & {[}1, 1{]}                                 & {[}0.81, 1{]}                 & {[}0.57, 0.81{]}  & {[}0.81, 1{]}                 & {[}0.81, 1{]}                 & {[}1, 1{]}                                 & {[}0.81, 1{]}                 \\
1950 & {[}1, 1{]}                                 & {[}0.81, 1{]}                 & {[}0.57, 0.81{]}  & {[}0.81, 1{]}                 & {[}0.81, 1{]}                 & {[}1, 1{]}                                 & {[}0.81, 1{]}                 \\
2000 & {[}1, 1{]}                                 & {[}0.81, 1{]}                 & {[}0.57, 0.81{]}  & {[}0.81, 1{]}                 & {[}0.81, 1{]}                 & {[}1, 1{]}                                 & {[}0.81, 1{]}                 \\
2050 & {[}1, 1{]}                                 & {[}0.81, 1{]}                 & {[}0.81, 0.81{]} & {[}0.81, 1{]}                 & {[}0.81, 1{]}                 & {[}1, 1{]}                                 & {[}0.81, 1{]}                 \\
2100 & {[}1, 1{]}                                 & {[}0.81, 1{]}                 & {[}0.81, 0.81{]} & {[}0.81, 1{]}                 & {[}0.81, 1{]}                 & {[}1, 1{]}                                 & {[}0.81, 1{]}                 \\
2150 & {[}1, 1{]}                                 & {[}0.81, 1{]}                 & {[}0.81, 0.81{]} & {[}0.81, 1{]}                 & {[}0.81, 1{]}                 & {[}1, 1{]}                                 & {[}0.81, 1{]}                 \\
2200 & {[}1, 1{]}                                 & {[}0.81, 1{]}                 & {[}0.81, 0.81{]} & {[}0.81, 1{]}                 & {[}0.81, 1{]}                 & {[}1, 1{]}                                 & {[}1, 1{]}                                 \\
2250 & {[}1, 1{]}                                 & {[}0.81, 1{]}                 & {[}0.81, 1{]}                 & {[}0.81, 1{]}                 & {[}0.81, 1{]}                 & {[}1, 1{]}                                 & {[}1, 1{]}                                 \\
2300 & {[}1, 1{]}                                 & {[}0.81, 1{]}                 & {[}0.81, 1{]}                 & {[}0.81, 1{]}                 & {[}0.81, 1{]}                 & {[}1, 1{]}                                 & {[}1, 1{]}                                 \\
2350 & {[}1, 1{]}                                 & {[}0.81, 1{]}                 & {[}0.81, 1{]}                 & {[}0.81, 1{]}                 & {[}0.81, 1{]}                 & {[}1, 1{]}                                 & {[}1, 1{]}                                 \\
2400 & {[}1, 1{]}                                 & {[}0.81, 1{]}                 & {[}0.81, 1{]}                 & {[}0.81, 1{]}                 & {[}0.81, 1{]}                 & {[}1, 1{]}                                 & {[}1, 1{]}                                 \\
2450 & {[}1, 1{]}                                 & {[}0.81, 1{]}                 & {[}0.81, 1{]}                 & {[}0.81, 1{]}                 & {[}0.81, 1{]}                 & {[}1, 1{]}                                 & {[}1, 1{]}                                 \\
2500 & {[}1, 1{]}                                 & {[}0.81, 1{]}                 & {[}0.81, 1{]}                 & {[}0.81, 1{]}                 & {[}0.81, 1{]}                 & {[}1, 1{]}                                 & {[}1, 1{]}                                 \\
2550 & {[}1, 1{]}                                 & {[}0.81, 1{]}                 & {[}0.81, 1{]}                 & {[}0.81, 1{]}                 & {[}1, 1{]}                                 & {[}1, 1{]}                                 & {[}1, 1{]}                                 \\
2600 & {[}1, 1{]}                                 & {[}1, 1{]}                                 & {[}0.81, 1{]}                 & {[}1, 1{]}                                 & {[}1, 1{]}                                 & {[}1, 1{]}                                 & {[}1, 1{]}                                 \\
2650 & {[}1, 1{]}                                 & {[}1, 1{]}                                 & {[}0.81, 1{]}                 & {[}1, 1{]}                                 & {[}1, 1{]}                                 & {[}1, 1{]}                                 & {[}1, 1{]}                                 \\
2700 & {[}1, 1{]}                                 & {[}1, 1{]}                                 & {[}0.81, 1{]}                 & {[}1, 1{]}                                 & {[}1, 1{]}                                 & {[}1, 1{]}                                 & {[}1, 1{]}                                 \\
2750 & {[}1, 1{]}                                 & {[}1, 1{]}                                 & {[}1, 1{]}                                 & {[}1, 1{]}                                 & {[}1, 1{]}                                 & {[}1, 1{]}                                 & {[}1, 1{]}                                 \\
2800 & {[}1, 1{]}                                 & {[}1, 1{]}                                 & {[}1, 1{]}                                 & {[}1, 1{]}                                 & {[}1, 1{]}                                 & {[}1, 1{]}                                 & {[}1, 1{]}                                
\end{tabular}

\label{ci}
\end{table*}

\begin{figure}[h!]
		\includegraphics[width=\columnwidth]{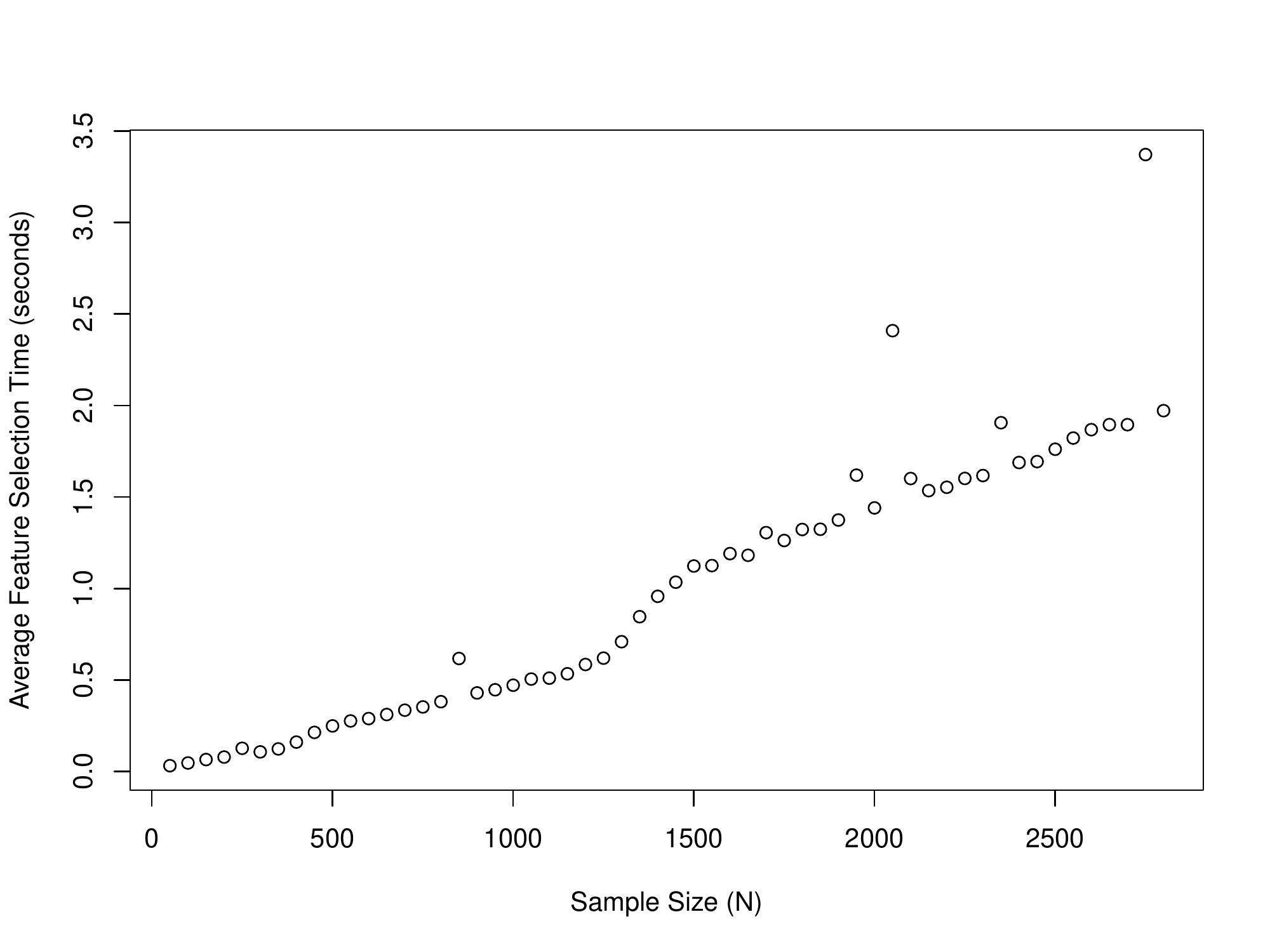}
		\caption{The average computation time of the proposed method when implementing feature selection in R.}
		\label{figure2}
	\end{figure}

	Meanwhile, we recorded the average computation time of the proposed method when implementing feature selection in R. The plot of results is shown in Figure \ref{figure2}. The computation time when $N=50$ was 0.03 seconds; the time when $N=2800$ was 1.97 seconds; the longest time during the simulation was 3.37 seconds.

	Based on the simulation result in Figure \ref{figure1}, different methods achieve 1 (in average IRRs) at different sample sizes. One should realize that the threshold of a sufficiently large sample greatly depends on the probability spaces of the underlying associated features and the outcome. The probability spaces of real datasets are generally significantly more complicated than that of the simulation. Consequently, in reality, particularly in health data, the majority of samples should be considered small; hence, the efficiency of a feature selection method is very important.
	
	The simulation codes are available at \cite{SimulationCode}. The proposed feature selection method using CASMI are implemented in the R package at \cite{RPackage}.
	
	\section{Discussion}
	
	\label{discussion}
	
	In this article, we have proposed a new entropic feature selection method based on CASMI. Compared to existing methods, the proposed method has two unique advantages: 1) it is very efficient as the joint of selected features provides the most relevant information compared to features selected by other methods, particularly when the sample size is relatively small, and 2) it automatically learns the number of features to be selected from data. The proposed method handles feature redundancy from the perspective of joint-distributions. Although we initially developed the proposed method for the domain-specific challenges in healthcare, the proposed method can be used in many other areas where there is an issue of limited sample.
	
	The proposed method is an entropic information-theoretic method. It aims at assisting data analytics on non-ordinal spaces. However, the proposed method can also be used on numerical data with an appropriate binning technique. Furthermore, using the proposed method on binned numerical data could discover different information as the entropic method looks at the data from a non-ordinal perspective.

	In detecting unhelpful associations (situation 2 and 3 features), we implement an adjustment from the sample coverage. The level of this adjustment can be modified by users. For example, users can replace the scoring function of the proposed method by CASMI* with a tuning parameter ($u$) as follows:
	\begin{equation*}
	\kappa ^*(X,Y)=\kappa(X,Y)\cdot(1-\pi_0(X))^u,
	\end{equation*}
	and estimate it by
	\begin{equation*}
	\hat{\kappa}^*(X,Y)=\hat{\kappa}_z(X,Y)\cdot (1-T_1(X))^u,
	\end{equation*}
	where $u$ is any fixed positive number. The $u$ can be considered as a parameter to determine the requirement for a feature to qualify situation 1. A larger $u$ stands for a heavier penalty from the sample coverage; hence, a feature needs to contain more real information to be categorized to situation 1. A smaller $u$ stands for a less penalty from the sample coverage; hence, a feature with less real information could be categorized to situation 1. However, users should be cautious when using a small $u$ because it may mistakenly classify an irrelevant feature (situation 3) to situation 1, and further exacerbates the issue of generalization. We suggest to begin the proposed feature selection method with $u=1$. After completing feature selection, if a user desires to select more or less features, the user could re-run the proposed method with a smaller or larger $u$, respectively, and keep modifying the value of $u$ until satisfactory.

	The proposed method only selects features but does not provide a classifier; however, to draw inferences on outcomes, a classifier is needed. To this end, additional techniques are required, such as machine learning (e.g., regressions and random forest). Into the future, it may be interesting to explore 1) methods that can distinguish features under situation 2 and 3 when the sample size is small; and 2) the possibilities of extending the proposed method to tree-based algorithms (e.g., random forest) to help determine which leaves and branches should be omitted. In addition, it may be interesting to investigate the performance of existing entropic methods if we use the $\hat{H}_z$, instead of $\hat{H}$, to estimate the entropies in their score functions.

	\section*{Acknowledgement}
	The authors of this article thank Prof. Zhiyi Zhang, at UNC Charlotte, for sharing his meaningful thoughts with us during this research.
	
	
	\bibliography{mybibfile}

\begin{thebibliography}{10}
\expandafter\ifx\csname url\endcsname\relax
  \def\url#1{\texttt{#1}}\fi
\expandafter\ifx\csname urlprefix\endcsname\relax\def\urlprefix{URL }\fi
\expandafter\ifx\csname href\endcsname\relax
  \def\href#1#2{#2} \def\path#1{#1}\fi

\bibitem{kruse2016challenges}
C.~S. Kruse, R.~Goswamy, Y.~Raval, S.~Marawi, Challenges and opportunities of
  big data in health care: a systematic review, JMIR medical informatics 4~(4).

\bibitem{lee2017medical}
C.~H. Lee, H.-J. Yoon, Medical big data: promise and challenges, Kidney
  research and clinical practice 36~(1) (2017) 3.

\bibitem{blum1997selection}
A.~L. Blum, P.~Langley, Selection of relevant features and examples in machine
  learning, Artificial intelligence 97~(1-2) (1997) 245--271.

\bibitem{kohavi1997wrappers}
R.~Kohavi, G.~H. John, Wrappers for feature subset selection, Artificial
  intelligence 97~(1-2) (1997) 273--324.

\bibitem{li2017feature}
J.~Li, K.~Cheng, S.~Wang, F.~Morstatter, R.~P. Trevino, J.~Tang, H.~Liu,
  Feature selection: A data perspective, ACM Computing Surveys (CSUR) 50~(6)
  (2017) 94.

\bibitem{urbanowicz2018relief}
R.~J. Urbanowicz, M.~Meeker, W.~La~Cava, R.~S. Olson, J.~H. Moore, Relief-based
  feature selection: introduction and review, Journal of biomedical
  informatics.

\bibitem{cai2018feature}
J.~Cai, J.~Luo, S.~Wang, S.~Yang, Feature selection in machine learning: A new
  perspective, Neurocomputing 300 (2018) 70--79.

\bibitem{shannon1948mathematical}
C.~E. Shannon, A mathematical theory of communication, Bell system technical
  journal 27~(3) (1948) 379--423.

\bibitem{duda2012pattern}
R.~O. Duda, P.~E. Hart, D.~G. Stork, Pattern classification, John Wiley \&
  Sons, 2012.

\bibitem{robnik2003theoretical}
M.~Robnik-{\v{S}}ikonja, I.~Kononenko, Theoretical and empirical analysis of
  relieff and rrelieff, Machine learning 53~(1-2) (2003) 23--69.

\bibitem{nie2008trace}
F.~Nie, S.~Xiang, Y.~Jia, C.~Zhang, S.~Yan, Trace ratio criterion for feature
  selection., in: AAAI, Vol.~2, 2008, pp. 671--676.

\bibitem{he2006laplacian}
X.~He, D.~Cai, P.~Niyogi, Laplacian score for feature selection, in: Advances
  in neural information processing systems, 2006, pp. 507--514.

\bibitem{zhao2007spectral}
Z.~Zhao, H.~Liu, Spectral feature selection for supervised and unsupervised
  learning, in: Proceedings of the 24th international conference on Machine
  learning, ACM, 2007, pp. 1151--1157.

\bibitem{liu2009slep}
J.~Liu, S.~Ji, J.~Ye, et~al., Slep: Sparse learning with efficient projections,
  Arizona State University 6~(491) (2009) 7.

\bibitem{nie2010efficient}
F.~Nie, H.~Huang, X.~Cai, C.~H. Ding, Efficient and robust feature selection
  via joint ℓ2, 1-norms minimization, in: Advances in neural information
  processing systems, 2010, pp. 1813--1821.

\bibitem{cai2010unsupervised}
D.~Cai, C.~Zhang, X.~He, Unsupervised feature selection for multi-cluster data,
  in: Proceedings of the 16th ACM SIGKDD international conference on Knowledge
  discovery and data mining, ACM, 2010, pp. 333--342.

\bibitem{yang2011robust}
F.~Yang, K.~Mao, Robust feature selection for microarray data based on
  multicriterion fusion, IEEE/ACM Transactions on Computational Biology and
  Bioinformatics 8~(4) (2011) 1080--1092.

\bibitem{li2012unsupervised}
Z.~Li, Y.~Yang, J.~Liu, X.~Zhou, H.~Lu, Unsupervised feature selection using
  nonnegative spectral analysis, in: Twenty-Sixth AAAI Conference on Artificial
  Intelligence, 2012.

\bibitem{davis1986statistics}
J.~C. Davis, R.~J. Sampson, Statistics and data analysis in geology, Vol. 646,
  Wiley New York et al., 1986.

\bibitem{tibshirani1996regression}
R.~Tibshirani, Regression shrinkage and selection via the lasso, Journal of the
  Royal Statistical Society: Series B (Methodological) 58~(1) (1996) 267--288.

\bibitem{lewis1992feature}
D.~D. Lewis, Feature selection and feature extraction for text categorization,
  in: Proceedings of the workshop on Speech and Natural Language, Association
  for Computational Linguistics, 1992, pp. 212--217.

\bibitem{battiti1994using}
R.~Battiti, Using mutual information for selecting features in supervised
  neural net learning, IEEE Transactions on neural networks 5~(4) (1994)
  537--550.

\bibitem{yang2000data}
H.~H. Yang, J.~Moody, Data visualization and feature selection: New algorithms
  for nongaussian data, in: Advances in Neural Information Processing Systems,
  2000, pp. 687--693.

\bibitem{vidal2003object}
M.~Vidal-Naquet, S.~Ullman, Object recognition with informative features and
  linear classification., in: IEEE Conference on Computer Vision and Pattern
  Recognition, Vol.~3, 2003, p. 281.

\bibitem{fleuret2004fast}
F.~Fleuret, Fast binary feature selection with conditional mutual information,
  Journal of Machine Learning Research 5~(Nov) (2004) 1531--1555.

\bibitem{peng2005feature}
H.~Peng, F.~Long, C.~Ding, Feature selection based on mutual information
  criteria of max-dependency, max-relevance, and min-redundancy, IEEE
  Transactions on pattern analysis and machine intelligence 27~(8) (2005)
  1226--1238.

\bibitem{lin2006conditional}
D.~Lin, X.~Tang, Conditional infomax learning: an integrated framework for
  feature extraction and fusion, in: European Conference on Computer Vision,
  Springer, 2006, pp. 68--82.

\bibitem{meyer2006use}
P.~E. Meyer, G.~Bontempi, On the use of variable complementarity for feature
  selection in cancer classification, in: Workshops on applications of
  evolutionary computation, Springer, 2006, pp. 91--102.

\bibitem{bennasar2015feature}
M.~Bennasar, Y.~Hicks, R.~Setchi, Feature selection using joint mutual
  information maximisation, Expert Systems with Applications 42~(22) (2015)
  8520--8532.

\bibitem{liu1995chi2}
H.~Liu, R.~Setiono, Chi2: Feature selection and discretization of numeric
  attributes, in: Proceedings of 7th IEEE International Conference on Tools
  with Artificial Intelligence, IEEE, 1995, pp. 388--391.

\bibitem{gini1912variabilita}
C.~Gini, Variabilita e mutabilita, studi economico-giuridici della r,
  Universita di Cagliari 3~(2) (1912) 3--159.

\bibitem{hall1999feature}
M.~A. Hall, L.~A. Smith, Feature selection for machine learning: comparing a
  correlation-based filter approach to the wrapper., in: FLAIRS conference,
  Vol. 1999, 1999, pp. 235--239.

\bibitem{harris1975statistical}
B.~Harris, The statistical estimation of entropy in the non-parametric case,
  Tech. rep., Wisconsin Univ-Madison Mathematics Research Center (1975).

\bibitem{paninski2003estimation}
L.~Paninski, Estimation of entropy and mutual information, Neural computation
  15~(6) (2003) 1191--1253.

\bibitem{good1953population}
I.~J. Good, The population frequencies of species and the estimation of
  population parameters, Biometrika 40~(3-4) (1953) 237--264.

\bibitem{kullback1951information}
S.~Kullback, R.~A. Leibler, On information and sufficiency, The annals of
  mathematical statistics 22~(1) (1951) 79--86.

\bibitem{wyner1978definition}
A.~D. Wyner, A definition of conditional mutual information for arbitrary
  ensembles, Information and Control 38~(1) (1978) 51--59.

\bibitem{guiasu1977information}
S.~Guiasu, Information theory with applications, Vol. 202, McGraw-Hill New
  York, 1977.

\bibitem{zhang2016statistical}
Z.~Zhang, Statistical Implications of Turing's Formula, John Wiley \& Sons,
  2016.

\bibitem{ohannessian2012rare}
M.~I. Ohannessian, M.~A. Dahleh, Rare probability estimation under regularly
  varying heavy tails, in: Conference on Learning Theory, 2012, pp. 21--1.

\bibitem{zhang2012entropy}
Z.~Zhang, Entropy estimation in turing's perspective, Neural computation 24~(5)
  (2012) 1368--1389.

\bibitem{zhang2015mutual}
Z.~Zhang, L.~Zheng, A mutual information estimator with exponentially decaying
  bias, Statistical applications in genetics and molecular biology 14~(3)
  (2015) 243--252.

\bibitem{dougherty1995supervised}
J.~Dougherty, R.~Kohavi, M.~Sahami, Supervised and unsupervised discretization
  of continuous features, in: Machine Learning Proceedings 1995, Elsevier,
  1995, pp. 194--202.

\bibitem{rubin1976inference}
D.~B. Rubin, Inference and missing data, Biometrika 63~(3) (1976) 581--592.

\bibitem{little2012prevention}
R.~J. Little, R.~D'agostino, M.~L. Cohen, K.~Dickersin, S.~S. Emerson, J.~T.
  Farrar, C.~Frangakis, J.~W. Hogan, G.~Molenberghs, S.~A. Murphy, et~al., The
  prevention and treatment of missing data in clinical trials, New England
  Journal of Medicine 367~(14) (2012) 1355--1360.

\bibitem{kang2013prevention}
H.~Kang, The prevention and handling of the missing data, Korean journal of
  anesthesiology 64~(5) (2013) 402--406.

\bibitem{zhang2018mutual}
J.~Zhang, C.~Chen, On `a mutual information estimator with exponentially
  decaying bias' by zhang and zheng, Statistical applications in genetics and
  molecular biology 17~(2).

\bibitem{pascoal2017theoretical}
C.~Pascoal, M.~R. Oliveira, A.~Pacheco, R.~Valadas, Theoretical evaluation of
  feature selection methods based on mutual information, Neurocomputing 226
  (2017) 168--181.

\bibitem{SimulationCode}
J.~Shi, Casmi simulation r codes,
  \url{https://github.com/JingyiShi/CASMI/blob/master/SimulationEvaluationUsingGroundTruth.R},
  [Online; Github] (2019).

\bibitem{RPackage}
J.~Shi, Casmi in r, \url{https://github.com/JingyiShi/CASMI}, [Online; Github]
  (2019).

\end{thebibliography}

\end{document}